\documentclass[10pt,twocolumn,letterpaper]{article}

\usepackage[pagenumbers]{cvpr} 

\definecolor{cvprblue}{rgb}{0.21,0.49,0.74}
\usepackage[pagebackref,breaklinks,colorlinks,allcolors=cvprblue]{hyperref}
\usepackage{algorithm}
\usepackage{algpseudocode} 
\usepackage{multirow}
\usepackage{tikz}
\usetikzlibrary{shapes,arrows,positioning,fit}
\usepackage{pgfplots}
\usepackage{pgfplotstable}
\usepackage{siunitx} 
\pgfplotsset{compat=1.18}

\usepgfplotslibrary{colorbrewer}
\pgfplotsset{cycle list/Set1-6} 
\definecolor{majorColor}{RGB}{31,119,180} 
\definecolor{minorColor}{RGB}{214,39,40}  
\usepackage{graphicx}
\usepackage{xcolor}
\usepackage{amsmath}
\usetikzlibrary{positioning, shapes.multipart, fit, calc, arrows.meta, decorations.pathreplacing}
\usepackage{graphicx} 
\usepackage{subcaption} 
\usepackage[acronym]{glossaries}
\makeglossaries

\usepackage{makecell}
\usepackage{colortbl}%

\def\paperID{16937} 
\def\confName{CVPR}
\def\confYear{2025}

\title{CEReBrO: Compact Encoder for Representations of Brain Oscillations Using Efficient Alternating Attention}
\author{
    Alexandru Dimofte\textsuperscript{1} \thanks{Glenn Anta Bucagu and Alexandru Dimofte contributed equally to this work.} \quad
    Glenn Anta Bucagu\textsuperscript{1} \quad
    Thorir Mar Ingolfsson\textsuperscript{2} \quad \\
    Xiaying Wang\textsuperscript{2} \quad 
    Andrea Cossettini\textsuperscript{2} \quad
    Luca Benini\textsuperscript{2} \quad
    Yawei Li\textsuperscript{2} \thanks{Corresonpence to \texttt{yawli@iis.ee.ethz.ch}.} \\
    \textsuperscript{1}University of Zurich \qquad
    \textsuperscript{2}Integrated Systems Laboratory, ETH Z{\"u}rich, Z{\"u}rich, Switzerland.
}

\begin{document}

\newacronym{ofa}{OFA}{Once-For-All}
\newacronym{simd}{SIMD}{Single Instruction, Multiple Data}
\newacronym{elu}{ELU}{Exponential Linear Unit}
\newacronym{relu}{ReLU}{Rectified Linear Unit}
\newacronym{rpr}{RPR}{Random Partition Relaxation}
\newacronym{mac}{MAC}{Multiply Accumulate}
\newacronym{dma}{DMA}{Direct Memory Access}
\newacronym{bmi}{BMI}{Brain--Machine Interface}
\newacronym{bci}{BCI}{Brain--Computer Interface}
\newacronym{smr}{SMR}{Sensory Motor Rythms}
\newacronym{svm}{SVM}{Support Vector Machine}
\newacronym{svd}{SVD}{Singular Value Decomposition}
\newacronym{evd}{EVD}{Eigendecomposition}
\newacronym{iir}{IIR}{Infinite Impulse Response}
\newacronym{fir}{FIR}{Finite Impulse Response}
\newacronym{fc}{FC}{Fabric Controller}
\newacronym{nn}{NN}{Neural Network}
\newacronym{mrc}{MRC}{Multiscale Riemannian Classifier}
\newacronym{flop}{FLOP}{Floating Point Operation}
\newacronym{sos}{SOS}{Second-Order Section}
\newacronym{ipc}{IPC}{Instructions per Cycle}
\newacronym{tcdm}{TCDM}{Tightly Coupled Data Memory}
\newacronym{fpu}{FPU}{Floating Point Unit}
\newacronym{fma}{FMA}{Fused Multiply Add}
\newacronym{alu}{ALU}{Arithmetic Logic Unit}
\newacronym{dsp}{DSP}{Digital Signal Processing}
\newacronym{gpu}{GPU}{Graphics Processing Unit}
\newacronym{soc}{SoC}{System-on-Chip}
\newacronym{mi}{MI}{Motor-Imagery}
\newacronym{csp}{CSP}{Commmon Spatial Patterns}
\newacronym{fbcsp}{FBCSP}{Filter-Bank \acrlong{csp}}
\newacronym{pulp}{PULP}{parallel ultra-low power}
\newacronym{soa}{SoA}{state-of-the-art}
\newacronym{bn}{BN}{Batch Normalization}
\newacronym{isa}{ISA}{Instruction Set Architecture}
\newacronym{ecg}{ECG}{Electrocardiogram}
\newacronym{mcu}{MCU}{microcontroller}
\newacronym{rnn}{RNN}{recurrent neural network}
\newacronym{cnn}{CNN}{convolutional neural network}
\newacronym{tcn}{TCN}{temporal convolutional network}
\newacronym{emu}{EMU}{epilepsy monitoring unit}
\newacronym{ml}{ML}{Machine Learning}
\newacronym{dl}{DL}{Deep Learning}
\newacronym{ai}{AI}{Artificial Intelligence}
\newacronym{tcp}{TCP}{Temporal Central Parasagittal}
\newacronym{loocv}{LOOCV}{Leave-One-Out Cross-Validation}
\newacronym{wfcv}{WFCV}{Walk-Forward Cross-Validation}
\newacronym{rwcv}{RWCV}{Rolling Window Cross-Validation}
\newacronym{iot}{IoT}{Internet of Things}
\newacronym{auc}{AUC}{Area Under the Receiver Operator Characteristic}
\newacronym{dwt}{DWT}{Discrete Wavelet Transform}
\newacronym{fft}{FFT}{Fast Fourier Transform}
\newacronym{tpot}{TPOT}{Tree-based Pipeline Optimization Tool}

\newacronym{tuar}{TUAR}{Temple University Artifact Corpus}
\newacronym{tuev}{TUEV}{Temple University Event Corpus}

\newacronym{bss}{BSS}{Blind Source Separation}
\newacronym{ica}{ICA}{Independent Component Analysis}
\newacronym{ic}{ICs}{Independent Components}
\newacronym{asr}{ASR}{Artifact Subspace Reconstruction}
\newacronym{pca}{PCA}{Principal Component Analysis}
\newacronym{gap}{GAP}{Global Average Pooling}
\newacronym{fcn}{FCN}{Fully Connected Networks}
\newacronym{mlp}{MLP}{Multi-Layer Perceptron}
\newacronym{nas}{NAS}{Neural Architectural Search}
\newacronym{fph}{FP/h}{False Positives per Hour}
\newacronym{bvp}{BVP}{Blood volume Pulse}
\newacronym{eda}{EDA}{Electrodermal Activity}
\newacronym{acc}{ACC}{Accelerometer}
\newacronym{cae}{CAE}{Convolutional Autoencoder}
\newacronym{sswce}{SSWCE}{Sensitivity-Specificity Weighted Cross-Entropy}
\newacronym{ce}{CE}{Cross-Entropy}
\newacronym{ppg}{PPG}{plethysmography}
\newacronym{asic}{ASIC}{Application-specific integrated circuit}

\newacronym{cerebro}{CEReBrO}{Compact Encoder for Representations of Brain Oscillations using efficient alternating attention}
\newacronym{eeg}{EEG}{Electroencephalograph}
\newacronym[longplural={Small EEG Foundation Models}]{sefm}{SEFM}{Small EEG Foundation Model}
\newacronym[longplural={Large EEG Foundation Models}]{lefm}{LEFM}{Large EEG Foundation Model}
\newacronym[longplural={Large Language Models}]{llm}{LLM}{Large Language Model}
\newacronym[longplural={Small Language Models}]{slm}{SLM}{Small Language Model}
\newacronym{mae}{MAE}{Masked Autoencoding}
\newacronym{tueg}{TUEG}{Temple University EEG Corpus}
\newacronym{ieeg}{iEEG}{Intracranial Electroencephalograph}

\maketitle
\begin{abstract}
Electroencephalograph (EEG) is a crucial tool for studying brain activity. Recently, self-supervised learning methods leveraging large unlabeled datasets have emerged as a potential solution to the scarcity of widely available annotated EEG data. However, current methods suffer from at least one of the following limitations: i) sub-optimal EEG signal modeling, ii) model sizes in the hundreds of millions of trainable parameters, and iii) reliance on private datasets and/or inconsistent public benchmarks, hindering reproducibility. To address these challenges, we introduce a \textbf{C}ompact \textbf{E}ncoder for \textbf{Re}presentations of \textbf{Br}ain \textbf{O}scillations using alternating attention (\textbf{CEReBrO}), a new small EEG foundation model. Our tokenization scheme represents EEG signals at a per-channel patch granularity. We propose an alternating attention mechanism that jointly models intra-channel temporal dynamics and inter-channel spatial correlations, achieving $2\times$ speed improvement with $6\times$ less memory required compared to standard self-attention. We present several model sizes ranging from 3.6 million to 85 million parameters. Pre-trained on over 20,000 hours of publicly available scalp EEG recordings with diverse channel configurations, our models set new benchmarks in emotion detection and seizure detection tasks, with competitive performance in anomaly classification and gait prediction. This validates our models' effectiveness. 
\end{abstract}    
\section{Introduction}
\label{sec:intro}

An \gls{eeg} is a fundamental tool for capturing the brain's electrical activity, playing a crucial role in neuroscience research and clinical diagnostics~\cite{eegfundamentals}. Its applications are extensive, encompassing disease diagnosis, medical monitoring, and brain-computer interfacing \cite{Roy2019}. Modeling \gls{eeg} signals is inherently challenging due to their non-linear, correlated, and non-stationary
nature \cite{Rashid2016}, rendering classical time series models insufficient \cite{Brockwell2016}. 
The scarcity of annotated \gls{eeg}  data exacerbates these challenges. In fact, high-quality annotations require expert medical professionals, making the process time-consuming, costly, and prone to human error \cite{Boudewyn2023, Diachenko2022, gurnani2022data, 8714995}. For example, prolonged monitoring is crucial for conditions like epilepsy, which require continuous tracking over several days, adding to the burden on annotating experts. Furthermore, artifacts such as ocular, cardiac, and muscular activities can mimic seizure patterns and often lead to false alarms and misinterpretations \cite{tatum_artifact_2011, schirrmeister_deep_2017, mathias2019artifacts}. 

To address the limited availability of labeled data, self-supervised representation learning methods, \ie, \gls{eeg}  foundation models, have emerged (see \cref{sec:related_works}). These foundation models are pre-trained on large corpora of unlabeled \gls{eeg}  signals and can generalize to various downstream tasks through fine-tuning. While promising, these models face significant limitations:
\textbf{1) Sub-optimal balance of spatial and temporal characteristics of \gls{eeg}  signals}.

\begin{figure*}[htbp]
    \centering
    \resizebox{0.8\textwidth}{!}{%
        \includegraphics{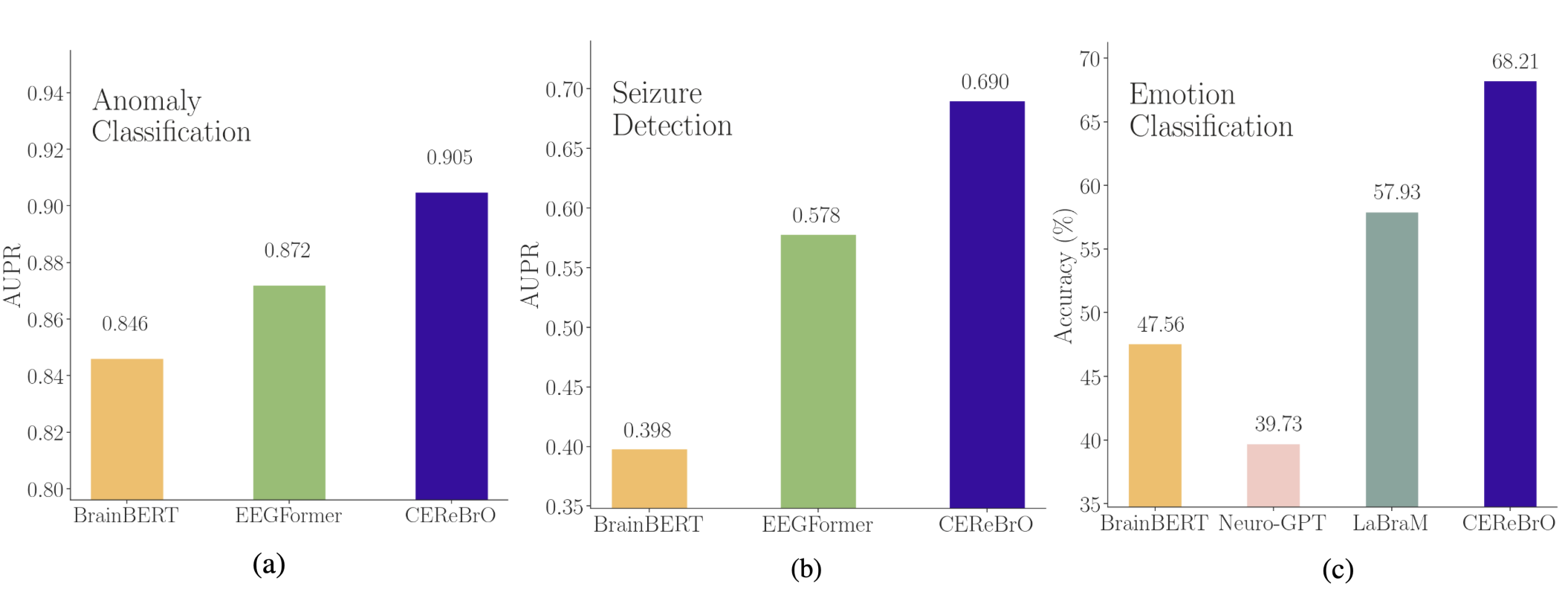} 
    }
    \vspace{-2mm}
    \caption{Comparison of CEReBrO with current SOTA in a) anomaly classification, b) seizure detection and c) emotion classification.}
    \vspace{-4mm}
    \label{fig:teaser_figure}
\end{figure*}

Self-supervised learning of \gls{eeg}  signals requires a careful balance between the temporal and spatial characteristics of \gls{eeg}  signals, especially in heterogeneous pre-training corpora with varying channel configurations. Some current models either ignore inter-channel spatial interactions \cite{EEGFormer, BrainBERT} or only consider auto-regressive temporal dynamics \cite{Neuro-GPT}. Other models strike a better balance between spatial and temporal characteristic modeling, but they either require multi-stage pre-training \cite{LaBraM} and/or architectures with hundreds of millions of trainable parameters \cite{Brant}.
\textbf{2) Increasing Model Sizes}. The trend toward \textit{\glspl{lefm}}, some exceeding hundreds of millions of parameters \cite{LaBraM, Brant, yuan2024brainwavebrainsignalfoundation}, raises concerns about computational efficiency and practicality. Without established scaling laws for \glspl{lefm} \cite{kaplan2020scalinglawsneurallanguage}, it is unclear if performance gains are due to model component design or model size. This is particularly problematic for real-time applications and deployment on resource-constrained devices like wearable \gls{eeg}  systems \cite{ingolfsson2024brainfusenet}. Developing a \gls{sefm} addresses these issues by enabling efficient, real-time processing suitable for widespread deployment. Without a universally accepted definition of \glspl{sefm}, we let edge device capabilities dictate our model sizes. The upper limit of tens of millions of trainable parameters is set by the latest smartphones \cite{apple2022deploying, liu2024mobilellmoptimizingsubbillionparameter} and edge TPU accelerators \cite{reidy2023efficient}.



\textbf{3) Data Inconsistency and Privacy}. The reliance on private datasets hinders reproducibility and standard benchmarking. Current methods vary in their use of private and public datasets for pre-training and fine-tuning. While some methods \cite{Neuro-GPT, EEGFormer} exclusively use public datasets, they measure downstream performance on different benchmarks. This complicates comparisons across different models.

Given these challenges, there is a pressing need for efficient \gls{eeg}  models that can handle complex signal dynamics, are suitable for deployment on limited hardware, and promote reproducibility through the use of public datasets. To overcome these limitations we introduce \textbf{\gls{cerebro}}, a \gls{sefm} featuring a novel tokenization scheme and an alternating attention mechanism. By alternating between modeling temporal dynamics within individual channels and spatial correlations across different channels, our method effectively captures the complex structure of \gls{eeg}  signals in a single encoder, accommodating diverse channel configurations found in different \gls{eeg}  devices. Our experiments show that this yields significant memory and runtime reductions with improvements in downstream tasks compared to standard self-attention. We pre-train \gls{cerebro} in three sizes (3.6M, 40M, and 85M parameters), drawing inspiration from \glspl{slm} \cite{jiao2020tinybertdistillingbertnatural, liu2024mobilellmoptimizingsubbillionparameter, lu2024smalllanguagemodelssurvey} to balance performance and efficiency. Utilizing the \gls{tueg} \cite{obeid2016temple} for pre-training ensures reproducibility and facilitates standardized benchmarking. Comprehensive evaluations on public datasets demonstrate \gls{cerebro}'s superior performance and generalizability. These efficiency gains make \gls{cerebro} particularly suitable for deployment on resource-constrained devices, facilitating real-time \gls{eeg}  analysis in wearable technology.

Our contributions are as follows:
\begin{itemize}
  \item \textbf{Novel Alternating Attention Mechanism}: We propose an efficient method for multi-channel \gls{eeg}  modeling that captures both temporal and spatial information, leading to significant memory reductions (up to $6\times$) and runtime reductions (up to $2\times$) compared to standard self-attention.
  \item \textbf{Development of \gls{cerebro}}: We introduce a compact \gls{sefm} that achieves competitive performance with significantly fewer parameters, making it suitable for deployment on resource-constrained devices and facilitating real-time \gls{eeg}  analysis.
  \item \textbf{Large-scale pre-training on \gls{tueg}}: We leverage over 20,000 hours of public \gls{eeg}  recordings from more than 10,000 unique subjects, encompassing diverse channel configurations, to ensure robustness and generalizability. To our knowledge, our method is the first to use token padding in pre-training to accommodate different channel configurations in \gls{tueg}.
  \item \textbf{Comprehensive Evaluations:} We validate \gls{cerebro} on multiple public benchmarks including anomaly classification, seizure detection, emotion recognition, gait prediction and achieve state-of-the-art performance. 
\end{itemize}

\section{Related Works} \label{sec:related_works}

\paragraph{\gls{eeg}  Foundation Models} Recent advancements in \gls{eeg}  foundation models have leveraged Transformer architectures and self-supervised learning techniques to address the scarcity of labeled \gls{eeg}  data. These models aim to learn robust representations from raw \gls{eeg}  signals, facilitating various downstream tasks without extensive labeled datasets.

BENDR \cite{bendr} is inspired by wav2vec \cite{wav2vec2} and employs a stack of short-receptive-field 1D convolutions to transform raw \gls{eeg}  waveforms into a sequence of embeddings, which are fed to a Transformer encoder with linear attention modules. Pre-training is done via a contrastive learning objective. However, without explicit channel-specific embeddings, BENDR may not optimally differentiate between channels, especially when the number of channels varies between training examples.
 
BrainBERT \cite{BrainBERT} uses \gls{ieeg}  spectrogram patches as input tokens to a standard Transformer encoder model, pre-trained via \gls{mae}. By processing channels individually, BrainBERT fails to capture inter-channel correlations, missing crucial spatial relationships inherent in multi-channel \gls{eeg} data. 

LaBraM \cite{LaBraM} introduces a learned neural tokenizer that maps patches of \gls{eeg}  waveforms to discrete codebook embeddings, which are then processed by a Transformer encoder within a symmetric \gls{mae} framework. The primary limitation of LaBraM is the additional computational overhead required to train the neural tokenizer.
Similarly, EEGFormer \cite{EEGFormer} incorporates a vector quantizer within a Transformer-based model trained via autoencoding, adding complexity to the training process. 

Neuro-GPT \cite{Neuro-GPT} adapts causal auto-regressive \gls{mae} for \gls{eeg}  waveform modeling. In this model, each token aggregates information from multiple channels, limiting the attention mechanism's ability to balance spatio-temporal characteristics. Neuro-GPT standardizes to 22 channels via nearest neighbor interpolation, which can degrade performance by losing detail from high channel counts and introducing artifacts when interpolating from fewer channels.

\paragraph{Vision Foundation Models} We draw on vision foundation models, especially Transformers and self-supervised learning, to enhance \gls{eeg}  signal modeling.

ChannelViT \cite{channelvit} modifies ViT's \cite{vit} patch projection, creating tokens for each channel-patch pair instead of aggregating channels. We adapt this approach from supervised training with RGB, satellite, and microscopy images to self-supervised learning with multi-channel \gls{eeg} , combining it with alternating attention to capture both temporal and spatial dynamics.

SimMIM \cite{simmim} and ViTMAE \cite{vitmae} offer distinct self-supervised learning approaches for vision models. SimMIM masks random tokens and processes all tokens through a Transformer encoder, using a linear layer for reconstruction. ViTMAE employs an encoder-decoder architecture, processing only visible tokens in the encoder and reconstructing the input in the decoder. Both optimize masked patch reconstruction. We adapt these \gls{mae} styles for \gls{eeg} , extending the loss function to better capture \gls{eeg}  characteristics.

Our approach addresses limitations in existing \gls{eeg}  foundation models by adapting vision model tokenization and masked image modeling. In addition, we introduce alternating attention, efficiently capturing temporal and spatial dynamics across diverse channel configurations while reducing computational demands for resource-constrained deployment.

\section{Methodology}
\label{sec:methodology}

\subsection{Pipeline Overview} In this section, we present \textbf{CEReBrO}, a novel \gls{sefm} designed to model \gls{eeg} waveforms using a compact encoder-only architecture efficiently. We introduce a novel tokenization scheme where \gls{eeg} waveforms are segmented into patches on a per-channel basis, allowing for granular modeling of temporal dynamics within each channel. This approach uniquely captures both intra-channel and inter-channel correlations, a technique not previously applied to \gls{eeg} signals to our knowledge. Our encoder consists of Transformer encoder blocks with a novel alternating attention mechanism. Alternating attention facilitates joint modeling of intra-channel temporal dynamics and inter-channel spatial correlations. During pre-training, we randomly mask input tokens. The encoder is optimized by reconstructing the original signals. Key innovations of our pipeline are discussed below.

\subsection{Tokenization} 
\label{subsec:tokenization}
Following current literature \cite{patchtst}, we slice \gls{eeg} waveforms into equally-sized non-overlapping patches to:
    i) enhance the locality and extract semantic information,
    ii) reduce computation and memory usage,
    iii) and attend to longer temporal dependencies.

Given an \gls{eeg} waveform $\mathbf{X} \in \mathbb{R}^{T \times C}$, where $T$ is the number of timestamps and $C$ is the number of channels, we segment $\mathbf{X}$ into non-overlapping patches of length $L$ with stride $S$. This results in a set of patches $\mathbf{P} \in \mathbb{R}^{N_p \times C \times L}$, where $N_p = \left\lfloor\frac{T-L}{S} + 1\right\rfloor$ is the number of patches per channel. Each patch $\mathbf{P}_{c,i} \in \mathbb{R}^{L}$ from channel $c$ and patch index $i$ is projected onto an embedding space of dimension $d_e$ using a learnable linear projection $\mathbf{W}_{\text{proj}} \in \mathbb{R}^{d_e \times L}$. The embedded patches are given by $\mathbf{E}_{c,i} = \mathbf{W}_\text{proj} \mathbf{P}_{c,i}^\top$. We then add learnable positional embeddings $\mathbf{W}_{\text{pos}} \in \mathbb{R}^{N_p \times d_e}$ and channel embeddings $\mathbf{W}_\text{chan} \in \mathbb{R}^{C \times d_e}$. The input embeddings are calculated as:
$$\mathbf{E}_{c,i}^{\text{in}} = \mathbf{E}_{c,i} + \textbf{W}_{\text{pos},i} + \textbf{W}_{\text{chan},c}$$ where $\mathbf{W}_{\text{pos},i} \in \mathbb{R}^{d_e}$ is the positional embedding for patch index $i$, and $\mathbf{W}_{\text{chan},c} \in \mathbb{R}^{d_e}$ is the channel embedding for channel $c$. This per-channel patch granularity of our tokens allows us to model intra-channel and inter-channel correlations jointly.

\begin{figure*}[t]
    \centering
    \includegraphics[width=\textwidth, keepaspectratio]{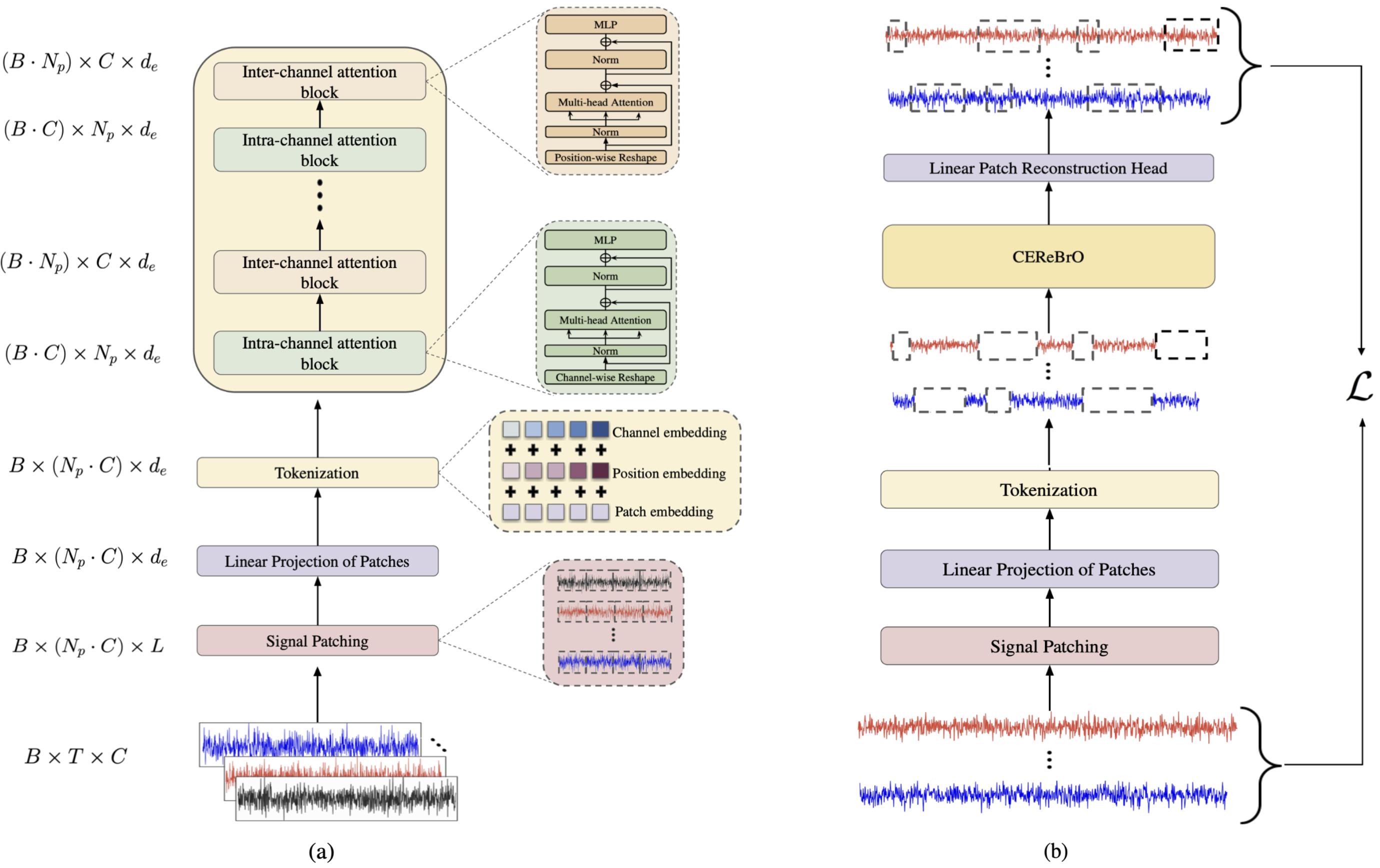}
    \vspace{-4mm}
    \caption{(a) Overview of the CEReBrO architecture. (b) Overview of our pre-training framework.}
    \vspace{-4mm}
    \label{fig:full_pipeline}
\end{figure*}

\subsection{Model Architecture} 
We use a Transformer encoder model \cite{vaswani2017attention} with $N$ layers, each comprising $H$ attention heads and embedding dimension $d_e$. Due to the fine-grained tokenization, each training example can contain up to several thousand tokens, increasing the computation complexity of the self-attention mechanism that scales quadratically with sequence length. To address this, we introduce an alternating attention mechanism (\cref{alg:alternating_attention}) within the Transformer encoder. This mechanism alternates between intra-channel and inter-channel attention in successive layers. 

In intra-channel attention layers, we focus on temporal dependencies within each channel by computing attention over the sequence of patches in that channel. In inter-channel attention layers, we capture spatial correlations by computing attention across channels at each time step. We use an even number of encoder layers so that both intra-channel and inter-channel attention are equally represented. The inputs to the first layer are the previously introduced input embeddings $\mathbf{E}_{c,i}^\text{in}$. Each layer alternates between intra-channel and inter-channel multi-head attention, applies a feed-forward network, and is followed by layer normalization. The outputs $\mathbf{E}_{c,i}^{(j)}$ of the $j$-th layer serve as the inputs to the $(j+1)$-th layer. 

During both pre-training and fine-tuning, the output of the final encoder layer is passed to a single linear layer. In pre-training, this linear layer serves as a patch reconstruction head; in fine-tuning, it functions as a classification layer. Our alternating attention mechanism is notable for two reasons: i) it allows us to jointly model intra-channel temporal dynamics and inter-channel spatial correlations within a single encoder, and ii) it enables efficient attention computation over long \gls{eeg} token sequences. We illustrate our end-to-end pipeline in \cref{fig:full_pipeline}.

\begin{algorithm}
\caption{\textproc{Alternating Attention Mechanism}}
\begin{algorithmic}[1]
\Require Input tensor $\textbf{T}$ of shape $[B, C \times N_p, d_e]$
\Statex \textbf{Parameters:}
\Statex \quad $B$: batch size
\Statex \quad $C$: number of channels
\Statex \quad $N_p$: number of patches per channel
\Statex \quad $d_e$: embedding dimension
\For{each encoder layer $i$ to $N$}
    \If{$i$ is odd} \Comment{Inter-channel attention}
        \State Reshape $\mathbf{T}$ to $[B \times N_p, C, d_e]$
        \State Compute QKV projection
        \State Multi-head attention over $C$ (channels)
    \Else \Comment{Intra-channel attention}
        \State Reshape $\mathbf{T}$ to $[B \times C, N_p, d_e]$
        \State Compute QKV projection
        \State Multi-head attention over $N_p$ (patches)
    \EndIf
    \State Reshape output back to $[B, C \times N_p, d_e]$
\EndFor
\Ensure Output tensor of shape $[B, C \times N_p, d_e]$
\end{algorithmic}
\label{alg:alternating_attention}
\end{algorithm}

\subsection{Theoretical and Empirical Analysis of Alternating Attention}
For standard self-attention, the memory complexity is quadratic with respect to sequence length because the attention mechanism considers pairwise interactions between all tokens in the sequence. In our alternating attention mechanism, the memory complexity alternates between:
\begin{itemize}
    \item \textbf{intra-channel attention} layers, where attention is computed within each channel independently over $N_p$ patches, resulting in a total complexity of $\mathcal{O}(C N_p^2)$.
    \item \textbf{inter-channel attention} layers, where attention is computed across $C$ channels at each of the $N_p$ time steps, resulting in a complexity of $\mathcal{O}(C^2 N_p)$.
\end{itemize}
Overall, the alternating attention mechanism significantly reduces the memory and computational requirements compared to standard self-attention, mainly when $C$ and $N_p$ are large. This reduction enables efficient processing of long \gls{eeg} sequences and high-channel-count data.
\begin{table}[!t]
    \centering
    \resizebox{\columnwidth}{!}{%
    \begin{tabular}{lll}
        \toprule
        \textbf{Attention Type} & \textbf{Memory Complexity} & \textbf{Time Complexity} \\
        \midrule
        Intra-channel & $\mathcal{O}\left( CN_p^2 \right)$ & $\mathcal{O}\left( CN_p^2d_e \right)$ \\
        Inter-channel & $\mathcal{O}\left( C^2N_p \right)$ & $\mathcal{O}\left( C^2N_pd_e \right)$ \\
        Standard Self-Attention & $\mathcal{O}\left( (CN_p)^2\right)$ & $\mathcal{O}\left( (CN_p)^2d_e \right)$ \\
        \bottomrule
    \end{tabular}%
    }
    \vspace{-2mm}
    \caption{Theoretical memory and time complexities of each attention type.}
    \label{tab:split_attn_complexity}
    \vspace{-4mm}
\end{table}

\begin{figure*}[ht]
  \centering
  \begin{subfigure}{0.45\textwidth}
    \centering
    \begin{tikzpicture}
      \begin{axis}[
        width=\textwidth,
        height=5cm,
        tick align=outside,
        tick pos=left,
        xmajorgrids,
        ymajorgrids,
        grid style={line width=.1pt, draw=gray!30},
        xlabel={Sequence Length},
        ylabel={GPU Memory Usage (MB)},
        xmin=0, xmax=4000,
        ymin=0, ymax=7000,
        xtick style={color=black},
        ytick style={color=black},
        tick label style={font=\normalsize},
      ]
        \addplot+[semithick, color=majorColor, solid, mark=triangle*, mark size=2]
        table {%
        64 33.6506881713867
        320 409.152008056641
        576 503.097869873047
        832 645.409301757812
        1088 839.756286621094
        1344 1083.84509277344
        1600 1377.28259277344
        1856 1722.81298828125
        2112 2117.60180664062
        2368 2562.72241210938
        2624 3059.15747070312
        2880 3604.67919921875
        3136 4200.53271484375
        3392 4846.92236328125
        3648 5543.537109375
        3904 6292.5810546875
        };
        \addplot+[semithick, color=majorColor, dashed, mark=*, mark size=2]
        table {%
        64 193.671173095703
        320 240.760314941406
        576 343.225860595703
        832 495.171081542969
        1088 699.086364746094
        1344 952.743408203125
        1600 1255.29040527344
        1856 1609.3486328125
        2112 2014.06616210938
        2368 2468.4599609375
        2624 2977.31420898438
        2880 3531.12622070312
        3136 4135.99072265625
        3392 4791.84228515625
        3648 5497.04248046875
        3904 6253.75390625
        };
        \addplot+[semithick, color=majorColor, dotted, mark=square*, mark size=2]
        table {%
        64 528.556030273438
        320 573.461486816406
        576 679.531494140625
        832 838.29248046875
        1088 1044.763671875
        1344 1304.45007324219
        1600 1611.32238769531
        1856 1972.58142089844
        2112 2379.953125
        2368 2840.80224609375
        2624 3348.83740234375
        2880 3910.34985351562
        3136 4519.04833984375
        3392 5181.2978515625
        3648 5890.8232421875
        3904 6653.826171875
        };
        \addplot+[semithick, color=minorColor, solid, mark=triangle*, mark size=2]
        table {%
        64 374.524932861328
        320 378.743286132812
        576 389.851654052734
        832 400.566772460938
        1088 411.937286376953
        1344 422.849029541016
        1600 433.564147949219
        1856 446.171661376953
        2112 456.657409667969
        2368 468.191741943359
        2624 479.660552978516
        2880 490.932739257812
        3136 501.156341552734
        3392 512.633361816406
        3648 524.356079101562
        3904 544.8115234375
        };
        \addplot+[semithick, color=minorColor, dashed, mark=*, mark size=2]
        table {%
        64 193.671173095703
        320 216.970748901367
        576 241.513977050781
        832 269.202941894531
        1088 295.581176757812
        1344 322.024963378906
        1600 345.978363037109
        1856 373.667327880859
        2112 398.210571289062
        2368 425.899505615234
        2624 453.391876220703
        2880 480.294403076172
        3136 504.051208496094
        3392 530.953735351562
        3648 554.907165527344
        3904 582.596069335938
        };
        \addplot+[semithick, color=minorColor, dotted, mark=square*, mark size=2]
        table {%
        64 528.556030273438
        320 554.849304199219
        576 589.616149902344
        832 625.431579589844
        1088 659.149841308594
        1344 694.703125
        1600 728.683532714844
        1856 765.146118164062
        2112 798.962707519531
        2368 834.352111816406
        2624 868.168701171875
        2880 903.558166503906
        3136 937.374694824219
        3392 972.837890625
        3648 1006.81829833984
        3904 1042.37158203125
        };
      \end{axis}
    \end{tikzpicture}
    \caption{}
    \label{fig:memory}
  \end{subfigure}
  \hfill
  \begin{subfigure}{0.45\textwidth}
    \centering
    \begin{tikzpicture}
      \begin{axis}[
        width=\textwidth,
        height=5cm,
        tick align=outside,
        tick pos=left,
        xmajorgrids,
        ymajorgrids,
        grid style={line width=.1pt, draw=gray!30},
        xlabel={Sequence Length},
        ylabel={Runtime (ms)},
        xmin=0, xmax=4000,
        ymin=0, ymax=800,
        xtick style={color=black},
        ytick style={color=black},
        tick label style={font=\normalsize},
      ]
      \addplot+[semithick, color=majorColor, solid, mark=triangle*, mark size=2]
      table {%
      64 14.9409532546997
      320 15.7184610366821
      576 19.7158279418945
      832 27.8810234069824
      1088 38.3830223083496
      1344 48.4120597839355
      1600 66.0469207763672
      1856 82.1626434326172
      2112 95.9837951660156
      2368 124.815689086914
      2624 160.964492797852
      2880 184.496475219727
      3136 221.638336181641
      3392 239.923202514648
      3648 278.453399658203
      3904 299.018646240234
      };
      \addplot+[semithick, color=majorColor, dashed, mark=*, mark size=2]
      table {%
      64 20.176549911499
      320 32.1553802490234
      576 45.3032455444336
      832 61.8463401794434
      1088 84.5827255249023
      1344 108.626235961914
      1600 137.85400390625
      1856 166.814163208008
      2112 194.224349975586
      2368 239.760772705078
      2624 296.272857666016
      2880 335.687652587891
      3136 395.931365966797
      3392 431.833557128906
      3648 495.899108886719
      3904 529.213989257812
      };
      \addplot+[semithick, color=majorColor, dotted, mark=square*, mark size=2]
      table {%
      64 38.2969093322754
      320 57.3398895263672
      576 82.5434646606445
      832 108.819458007812
      1088 141.172164916992
      1344 178.474029541016
      1600 222.633331298828
      1856 269.096374511719
      2112 310.054901123047
      2368 376.276733398438
      2624 452.013946533203
      2880 506.680145263672
      3136 585.4892578125
      3392 644.015563964844
      3648 731.463134765625
      3904 782.550415039062
      };
      \addplot+[semithick, color=minorColor, solid, mark=triangle*, mark size=2]
      table {%
      64 15.7883234024048
      320 16.2141208648682
      576 16.5049114227295
      832 17.1060314178467
      1088 19.5723571777344
      1344 19.7338466644287
      1600 22.3519687652588
      1856 22.7091941833496
      2112 25.1509571075439
      2368 27.5745429992676
      2624 29.3078231811523
      2880 30.5759143829346
      3136 31.5526523590088
      3392 33.8043556213379
      3648 35.3927192687988
      3904 37.6773681640625
      };
      \addplot+[semithick, color=minorColor, dashed, mark=*, mark size=2]
      table {%
      64 20.5178966522217
      320 30.2666301727295
      576 38.567081451416
      832 47.905330657959
      1088 57.2869834899902
      1344 68.6759948730469
      1600 79.8447418212891
      1856 86.6397171020508
      2112 96.5040817260742
      2368 107.179748535156
      2624 117.283744812012
      2880 126.524513244629
      3136 139.441009521484
      3392 149.735656738281
      3648 160.551025390625
      3904 166.976867675781
      };
      \addplot+[semithick, color=minorColor, dotted, mark=square*, mark size=2]
      table {%
      64 38.0889587402344
      320 55.147777557373
      576 73.9176025390625
      832 92.6613616943359
      1088 108.257614135742
      1344 128.539764404297
      1600 147.503768920898
      1856 167.431335449219
      2112 183.480606079102
      2368 203.755279541016
      2624 222.784515380859
      2880 238.563430786133
      3136 258.468048095703
      3392 280.417724609375
      3648 301.503082275391
      3904 315.707489013672
      };
      \end{axis}
    \end{tikzpicture}
    \caption{}
    \label{fig:runtime}
  \end{subfigure}
  \vspace{0.1em}
  \begin{tikzpicture}
    \node[draw=black, rounded corners, inner sep=2pt] (legend) {
      \begin{tikzpicture}[font=\normalsize]
        \matrix [column sep=4pt, row sep=2pt] {
          \draw[semithick, color=majorColor, solid, mark=triangle*, mark options={solid}, mark size=2]
            plot coordinates {(0,0)}; &
          \node[anchor=west] {Standard - Small}; &
          \draw[semithick, color=majorColor, dashed, mark=*, mark options={solid}, mark size=2]
            plot coordinates {(0,0)}; &
          \node[anchor=west] {Standard - Base}; &
          \draw[semithick, color=majorColor, dotted, mark=square*, mark options={solid}, mark size=2]
            plot coordinates {(0,0)}; &
          \node[anchor=west] {Standard - Large}; \\
          \draw[semithick, color=minorColor, solid, mark=triangle*, mark options={solid}, mark size=2]
            plot coordinates {(0,0)}; &
          \node[anchor=west] {Alternating - Small}; &
          \draw[semithick, color=minorColor, dashed, mark=*, mark options={solid}, mark size=2]
            plot coordinates {(0,0)}; &
          \node[anchor=west] {Alternating - Base}; &
          \draw[semithick, color=minorColor, dotted, mark=square*, mark options={solid}, mark size=2]
            plot coordinates {(0,0)}; &
          \node[anchor=west] {Alternating - Large}; \\
        };
      \end{tikzpicture}
    };
  \end{tikzpicture}

  \caption{Forward pass GPU Memory Usage (a) and Runtime (b) vs. Sequence Length for alternating attention and standard self-attention in three CEReBrO model sizes. We use $N_p = 20$ and $C \in [1, 64]$ to simulate typical \gls{eeg} configurations.}
   \vspace{-4mm}
  \label{fig:combined}
\end{figure*}
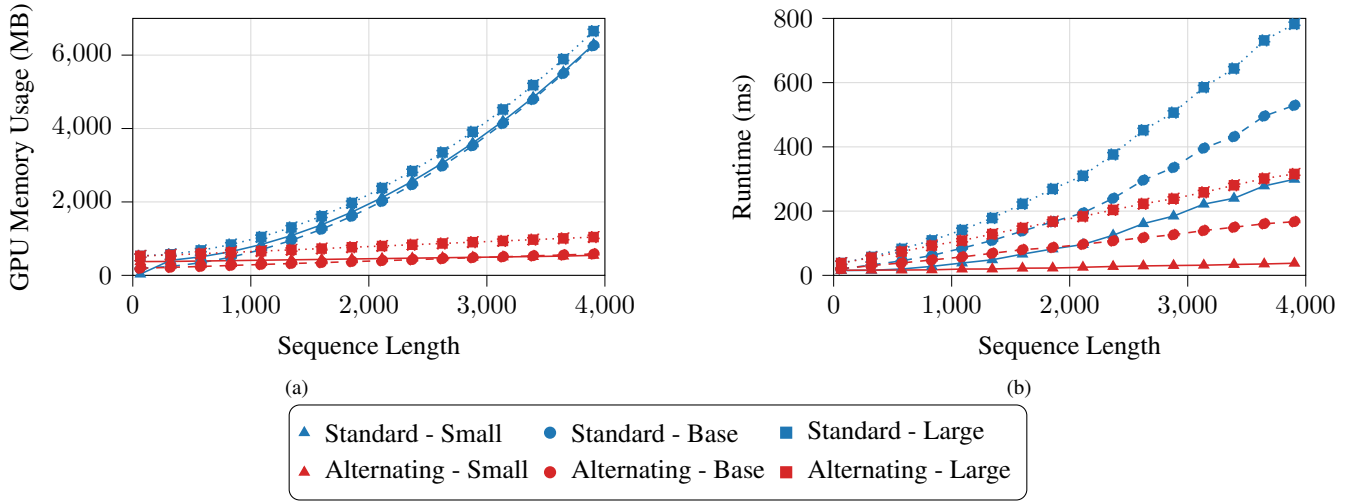

We empirically compare alternating attention and standard self-attention by measuring GPU memory requirements and runtime vs. sequence length of the forward pass for three different CEReBrO model sizes. To replicate conditions typically used in state-of-the-art models, we use $N_p = 64$ and $C \in [1, 64]$. Experiments are conducted on 4 NVIDIA RTX 2080 Ti GPUs. In \cref{fig:runtime}, we see that for the largest sequence lengths, assuming equal model size, standard self-attention yields $> 2\times$ the runtime compared to alternating attention. In \cref{fig:memory}, assuming equal model size, standard self-attention consumes $> 6\times$ the GPU memory consumed by alternating attention. These results are consistent with our theoretical analysis in \cref{tab:split_attn_complexity}. Alternating attention is particularly efficient for large values of $CN_p$. This includes settings with long \gls{eeg} signals and/or many \gls{eeg} channels. More details regarding split attention can be found in \cref{appendix:split_attention}.


There exist efficient alternatives to self-attention, including low-rank approximation methods \cite{chen2021scatterbrainunifyingsparselowrank}, linear attention \cite{katharopoulos2020transformersrnns} and sparse attention \cite{lou2024sparserfastermoreefficient} among many others. These methods typically provide memory and/or runtime improvements at the cost of performance compared to standard self-attention. This appears to be true for alternating attention applied to \gls{eeg} spectrograms. However, we find that alternating attention on \gls{eeg} waveforms can actually provide performance improvements compared to standard self-attention (see \cref{sec:experiments}). In theory, alternating attention can be combined with other self-attention alternatives (e.g. low rank approximations, linear attention etc) for potentially greater efficiency. Alternating attention can also, by design, be extended to general multi-channel timeseries and multi-channel images (although performance may vary by task). We leave these explorations for future research. 

\subsection{Pre-training} 
We use \gls{mae} during pre-training to allow our model to learn useful representations from a large unlabeled pre-training corpus. After patching of \gls{eeg} waveforms, we randomly mask a fixed portion of patches in each input sequence of patches $\mathbf{P}$. Masked positions are replaced with a single shared learnable [MASK] token. The resulting sequence of tokens is passed to our Transformer encoder. The encoder output is then passed to a linear layer that outputs a sequence $\hat{\mathbf{P}}$, which is a reconstruction of $\mathbf{P}$. We define the following loss components:

\begin{equation}
\mathcal{L_\text{masked}} = \frac{1}{|M|} \sum_{(c,i) \in M} \| \mathbf{P}_{c,i} - \hat{\mathbf{P}}_{c,i} \|_2^2
\end{equation}

\begin{equation}
\mathcal{L_\text{visible}} = \frac{1}{|\overline{M}|} \sum_{(c,i) \in \overline{M}} \| \mathbf{P}_{c,i} - \hat{\mathbf{P}}_{c,i} \|_2^2
\end{equation}
where $M$ and $\overline{M}$ are the set of masked and visible token positions respectively. Our total loss function during pre-training is:

\begin{equation}
\mathcal{L} = \mathcal{L_\text{masked}} + \alpha{\mathcal{L_\text{visible}}}
\end{equation}

Traditional \gls{mae} methods \cite{vitmae, audiomae, simmim, BrainBERT, bert} compute the reconstruction loss only on masked patches (i.e., $\alpha = 0$). We observed that doing so with our framework led to high-quality reconstructions of masked patches but relatively poor and unpredictable reconstructions of visible patches. Inspired by findings in\cite{vitmae}, which showed that computing the loss uniformly on all tokens (i.e., $\alpha = 1$) can reduce downstream performance, we introduce $\mathcal{L}_{\text{visible}}$ with a small weighting factor $\alpha = 0.1$. This approach stabilizes the reconstruction of visible patches while preventing the model from learning the identity function, thereby enhancing downstream performance. Similar strategies have been employed in lightweight vision transformers \cite{tan2024pretraininglightweightvisiontransformers}, demonstrating improved performance on low-resolution image datasets. We illustrate our pre-training framework in \cref{fig:full_pipeline}.

\subsection{Fine-tuning} 
We use our pre-trained encoder model as a feature extractor for a single linear layer classifier. While most \gls{mae} methods use a [CLS] token, we adopt global mean pooling over the token embeddings to aggregate information. This choice avoids additional computational overhead associated with a [CLS] token, which would require extra processing steps in the alternating attention mechanism framework. Empirically, using mean pooling yields comparable performance to using a [CLS] token \cite{vitmae}. The encoder weights can either be frozen (linear probing) or used as initialization (full fine-tuning). While full fine-tuning requires more computational resources, it generally yields better downstream performance. 

\subsection{Handling Different Channel Numbers In Pre-training} 
To handle varying numbers of channels, we pad the input sequences to a maximum channel count using a shared learnable [PAD] token, taking inspiration from LLMs~\cite{bert}. This allows the model to process inputs with different channel configurations within a unified architecture. The attention scores corresponding to [PAD] tokens are forced to zero during attention computations to prevent them from influencing the model's output. While padding increases the input sequence length, we limit the maximum number of channels to a reasonable upper bound (e.g. 64 channels) to control computational costs. This limit on the number of channels (and hence the context length of our model) justifies why padding is a reasonable mechanism for our specific case, as \gls{eeg} recording devices have a practical maximum number of channels due to physical constraints \cite{EGI_GSN}.

\section{Experiments}  \label{sec:experiments}

\subsection{Data} 
We pre-train our model on \gls{tueg} \cite{obeid2016temple}. This totals over 20,000 hours of \gls{eeg} signals from over 10,000 unique subjects and a wide variety of \gls{eeg} channel configurations (18-36 channels). This is the largest publicly available scalp \gls{eeg} corpus in the world and it covers an expansive range of pathologies and neural signals. We evaluate our model on several downstream tasks described by the following datasets: 
\begin{enumerate}[label=\roman*.]
\item TUAB \cite{obeid2016temple}: \gls{eeg} recordings (23 channels sampled at 250 Hz) from 2,329 subjects with normal/abnormal annotations.
\item SEED \cite{zheng2015investigating}: \gls{eeg} recordings (62 channels sampled at 200 Hz) from 15 subjects with annotations negative/neutral/positive emotion annotations in response to various movie clips. 
\item Neonate \cite{stevenson2019dataset}: \gls{eeg} recordings (19 channels sampled at 250 Hz) with annotated seizures for 79 Neonates.
\end{enumerate} 
All our pre-training and downstream data is publicly available, which we believe will facilitate benchmarking and development of \gls{eeg} foundation models. More details regarding our dataset pre-processing are available in \cref{appendix:experiment_details}.

\subsection{\gls{eeg} Representation and Pre-training Style} \label{subsec:waveform_vs_spectrograms}
We compare pre-training and fine-tuning with waveforms and spectrograms. While waveforms purely provide time information, spectrograms are a popular representation method in LEFMs \cite{BrainBERT} since they strike a balance between time and frequency resolution \cite{Feichtinger1998GaborAA}. Following \cite{BrainBERT}, we compute spectrogram representations using the Short-Time Fourier Transform. We also compare SimMIM and ViTMAE pre-training styles. While SimMiM uses a single-encoder only pre-training, ViTMAE separates pre-training into an encoder and lightweight Transformer decoder. In this setting, the encoder only provides representations of visible tokens and the decoder uses encoder outputs and masked tokens to reconstruct signals. For each candidate model, we fine-tune on TUAB and report balanced accuracy, macro AUPR and macro AUROC scores. We choose TUAB as our performance proxy because of its popularity in \gls{eeg} Foundation Model literature. To illustrate the importance of pre-training, we also show model performance when trained from scratch directly on TUAB. Our results are summarized in \cref{tab:representation_comparison}. More experiments on spectrograms can be found in \cref{appendix:spectrogram_ablations}.
\begin{table}[!ht]
    \centering
    \resizebox{\columnwidth}{!}{%
        \begin{tabular}{llccc}
            \toprule
            \textbf{Representation} & \textbf{Pre-training Style} & \textbf{Balanced Accuracy (\%)} & \textbf{AUPR} & \textbf{AUROC} \\
            \midrule
            \multirow{3}{*}{Spectrograms} & None & 76.49 & 0.8462 & 0.8483 \\
                                          & ViTMAE & 77.71 & 0.8586 & 0.8559 \\
                                          & SimMIM & 78.33 & 0.8659 & 0.8595 \\
            \midrule
            \multirow{3}{*}{Waveforms} & None & 76.39 & 0.8430 & 0.8426 \\
                                       & SimMIM & 79.35 & \textbf{0.8755} & 0.8700  \\
                                       & ViTMAE & \textbf{79.42} & 0.8741 & \textbf{0.8715} \\
            \bottomrule
        \end{tabular}%
    }
    \caption{Comparing \gls{eeg} signal representation and pre-training style.}
    \label{tab:representation_comparison}
\end{table}

From \cref{tab:representation_comparison}, we observe that pre-training significantly improves performance over training from scratch, with waveform representations slightly outperforming spectrograms in our setting. This suggests that preserving temporal information in raw waveforms may be more beneficial for certain \gls{eeg} tasks, possibly due to the loss of fine-grained temporal details in spectrograms. The performance gains, although modest compared to the amount of pre-training data used, align with observations in other works \citep{vitmae, simmim}, indicating diminishing returns with large-scale pre-training. Comparing pre-training styles, SimMIM and ViTMAE yield similar performance, with SimMIM being slightly better for spectrograms and ViTMAE for waveforms. This may be attributed to the inherent differences in how each method handles masked tokens and reconstruction objectives, affecting their ability to capture relevant features in different representations.

\subsection{Spectrogram Patching} 

\label{subsec:patching_experiments}
\begin{figure}[!ht]
    \centering
    \includegraphics[width=\columnwidth]{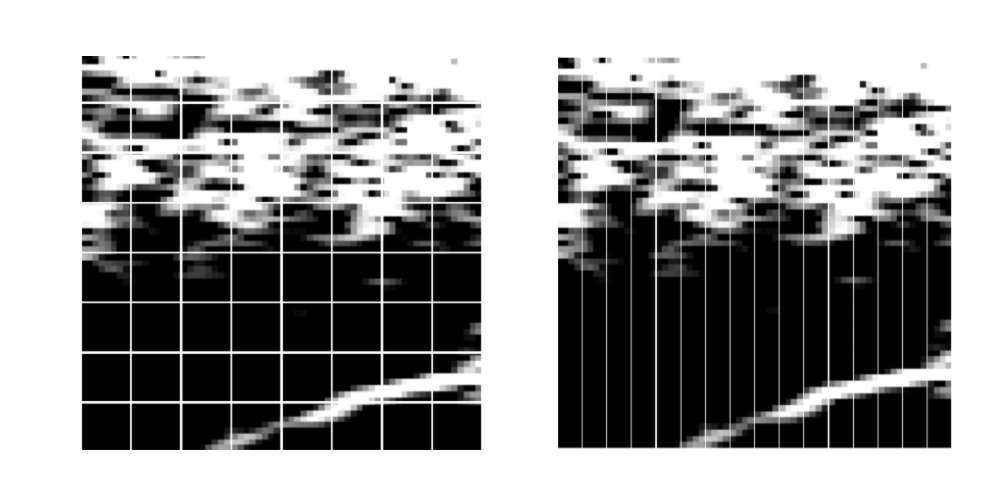}
    \caption{Spectrogram square patching (left) and time bin patching (right)}
    \label{fig:spectrogram_patching}
\end{figure}

Spectrograms can be patched in different ways. Square patches are typically used in vision foundation models \cite{vitmae, simmim}. Time bin patches have also been proposed specifically for spectrogram-related tasks \cite{BrainBERT, audiomae}. We illustrate the difference in \cref{fig:spectrogram_patching}.

We compare the use of square patches and time bin patches in spectrogram tokenization. Note that this also means we are comparing random unstructured masking with time bin masking. Results are outlined in \cref{tab:patching_comparison}.

\begin{table}[!ht]
    \centering
    \resizebox{\columnwidth}{!}{%
        \begin{tabular}{llccc}
            \toprule
            \textbf{Patching} & \textbf{Pre-training Style} & \textbf{Balanced Accuracy (\%)} & \textbf{AUPR} & \textbf{AUROC} \\
            \midrule
            Square patches & None & 76.49 & 0.8462 & 0.8483 \\
            Time bin patches & None & 77.24 & 0.8489 & 0.8498 \\
            \midrule
            Square patches & SimMIM & 78.33 & 0.8659 & 0.8595 \\
            Time bin patches & SimMIM & 78.89 & 0.8719 & 0.8690 \\
            \bottomrule
        \end{tabular}%
    }
    \caption{The effect of patch shape and masking strategy on downstream performance.}
    \label{tab:patching_comparison}
\end{table}

In \cref{tab:patching_comparison}, we observe that time bin patching yields better downstream performance on TUAB compared to the classical square patching. Time bin patching is both superior in the training from scratch and pre-trained regimes. We posit that this is partly due to the natural sequential nature of tokens when time bin patching is used. \\

\subsection{Alternating Attention vs. Standard Attention} \label{subsec:attention_experiments}
We aim to evaluate the impact of the alternating attention mechanism on downstream performance compared to using standard self-attention. Using the best-performing waveform and spectrogram models identified in \cref{subsec:waveform_vs_spectrograms}, we compare the two attention mechanisms. We focus on the SimMIM pre-training style, as the ViTMAE style benefits less from alternating attention due to its use of only visible tokens in the encoder. Our results are presented in \cref{tab:attention_comparison}. 

\begin{table}[!ht]
    \centering
    \resizebox{\columnwidth}{!}{%
        \begin{tabular}{lccccc}
            \toprule
            \textbf{Representation} & \textbf{Attention} & \textbf{Balanced Accuracy (\%)} & \textbf{AUPR} & \textbf{AUROC} \\
            \midrule
            \multirow{2}{*}{Spectrogram} 
             & Alternating & 78.49 & 0.8684 & 0.8674 \\
            & Self-Attention    & 78.89 & 0.8719 & 0.8690  \\
            \midrule
            \multirow{2}{*}{Waveform} 
            & Self-Attention    & 79.35 & 0.8755 & 0.8700  \\
            & Alternating & \textbf{80.65} & \textbf{0.8901} & \textbf{0.8825} \\
            \bottomrule
        \end{tabular}%
    }
    \caption{Comparing standard self-attention and alternating attention.}
    \label{tab:attention_comparison}
    \vspace{-2mm}
\end{table}

From \cref{tab:attention_comparison}, we observe differing outcomes between spectrograms and waveform representations. For spectrograms, standard self-attention marginally outperforms alternating attention across all performance metrics. This may be due to spectrograms already encapsulating frequency and temporal information in a two-dimensional format, where global attention can effectively capture relevant features.

In contrast, for waveforms, alternating attention significantly outperforms standard self-attention, achieving improvements of $+1.3\%$ in balanced accuracy, $+0.0146$ in AUPR and $+0.0125$ AUROC. This suggests that the alternating attention mechanism is more adept at capturing the inherent temporal dynamics and spatial correlations in raw waveforms, likely because it explicitly models intra-channel and inter-channel relationships. These results indicate that alternating attention is particularly beneficial when dealing with raw waveform data, emphasizing its suitability for modeling \gls{eeg} signals in their original temporal form.

\subsection{Comparisons with the State of the Art} \label{subsec:sota_comparisons}

We validate our model architecture and demonstrate its effectiveness in learning meaningful representations that generalize across various downstream tasks, including anomaly classification, emotion recognition and seizure detection\footnote{All performance metrics are established by existing methods. We are unable to train models other than \gls{cerebro} due to limited resources. "-" indicates missing information not found in publications, lacking open-source implementation, and for which the authors did not respond.}. Details regarding our experimental setup are in \cref{appendix:experiment_details}. We provide results on gait prediction in \cref{appendix:additional_downstream_tasks}.

\begin{table}[!ht]
    \centering
    \resizebox{\columnwidth}{!}{%
        \begin{tabular}{lcccc}
            \toprule
            \textbf{Method} & \textbf{Model Size} & \textbf{Accuracy (\%)} & \textbf{F1} \\
            \midrule
            \textbf{Supervised Models} \\
            LSTM  \cite{LSTM}       & -     & 44.31 & 0.4277 \\
            ConvNeXt  \cite{ConvNeXt}   & 3.7M  & 52.50 & 0.5155 \\
            \midrule
            \textbf{Self-supervised Models} \\
            BrainBERT \cite{BrainBERT}   & 43.18M    & 47.56 & 0.4679 \\
            Neuro-GPT \cite{Neuro-GPT}   & 79.53M    & 39.73 & 0.3949 \\
            LaBraM \cite{LaBraM}      & 46M & 57.93 & 0.5899 \\
            \midrule
            \textbf{CEReBrO}      & 3.58M     & 54.57  $\pm$ 0.56 & 0.5475 $\pm$ 0.0058 \\
            \textbf{CEReBrO}      & 39.95M     & 67.18 $\pm$ 0.24 & 0.6732 $\pm$ 0.0023 \\
            \textbf{CEReBrO}   & 85.15M    & \textbf{68.21}$\pm$ 0.63 & \textbf{0.6845} $\pm$ 0.0061 \\
            \bottomrule
        \end{tabular}%
    }
    \caption{Performance on emotion classification using SEED.}
    \vspace{-4mm}
    \label{tab:seed_comparison}
\end{table}

\cref{tab:seed_comparison} compares our approach to state-of-the-art emotion classification. Among the supervised methods, ConvNeXt achieves the highest performance while using 3.5M parameters. In comparison, BrainBERT and Neuro-GPT use more $ 11\times$ and $21\times$ more parameters respectively, while achieving lower performance. LaBraM remains competitive with $+2\%$ accuracy and $+0.3$ F1 score compared to \gls{cerebro} at a comparable size. At its smallest size, \gls{cerebro} outperforms all supervised approaches despite having less parameters, demonstrating the effectiveness of our method. At its largest size, \gls{cerebro} achieves a new state-of-the-art performance.

\begin{table}[!ht]
    \centering
    \resizebox{0.8\columnwidth}{!}{%
    \begin{tabular}{lccc}
        \hline
        \textbf{Method} & \textbf{Model Size} & \textbf{AUPR} & \textbf{AUROC} \\
        \hline
        \multicolumn{4}{l}{\textbf{Supervised Models}} \\
        EEGNet \cite{EEGNet} & - & 0.499 $\pm$ 0.044 & 0.793 $\pm$ 0.019 \\
        TCN \cite{TCN}  & - & 0.398 $\pm$ 0.025 & 0.731 $\pm$ 0.020 \\
        EEG-GNN \cite{EEG-GNN}  & 0.107M & 0.419 $\pm$ 0.021 & 0.760 $\pm$ 0.010 \\
        GraphS4mer \cite{graphs4mer}  & 0.266M & 0.374 $\pm$ 0.013 & 0.719 $\pm$ 0.007 \\
        STATENET \cite{STATENET} & - & \textbf{0.789} & \textbf{0.910} \\
        \hline
        \multicolumn{4}{l}{\textbf{Self-supervised Models}} \\
        BrainBERT \cite{BrainBERT} & 43.18M & 0.398 $\pm$ 0.027 & 0.734 $\pm$ 0.019 \\
        EEGFormer Small \cite{EEGFormer} & - & 0.578 $\pm$ 0.023 & 0.842 $\pm$ 0.008 \\
        EEGFormer Base \cite{EEGFormer} & - & 0.568 $\pm$ 0.036 & 0.842 $\pm$ 0.014 \\
        EEGFormer Large \cite{EEGFormer} & - & 0.544 $\pm$ 0.026 & 0.833 $\pm$ 0.017 \\
        \hline
        \textbf{CEReBrO} & 3.58M & 0.576 $\pm$ 0.089 & 0.830 $\pm$ 0.037 \\
        \textbf{CEReBrO} & 39.95M & 0.676 $\pm$ 0.092 & 0.867 $\pm$ 0.039\\
        \textbf{CEReBrO} & 85.15M & 0.690 $\pm$ 0.091 & 0.875 $\pm$ 0.039  \\
        \hline
    \end{tabular}%
    }
    \vspace{-2mm}
    \caption{Performance on seizure detection using the Neonate Seizure dataset}
    \vspace{-2mm}
    \label{tab:neonate_comparison}
\end{table}

\cref{tab:neonate_comparison} evaluates \textbf{\gls{cerebro}} in seizure detection. Our smallest model achieves a comparable performance to all variants of EEGFormer. When a larger model is considered (85M), we achieve a much higher AUROC, reaching a maximum value of 0.875.
While this number is still lower than the 0.91 from STATENET, \textbf{\gls{cerebro}} sets a new state-of-the-art among foundation models capable of handling different downstream tasks.

\begin{table}[!ht]
    \centering
    \setlength{\tabcolsep}{1.2mm}
    \resizebox{\columnwidth}{!}{%
        \begin{tabular}{lcccc}
            \toprule
            \textbf{Methods} & \textbf{Model Size} & \makecell{\textbf{Balanced} \\ \textbf{Accuracy (\%)}} & \textbf{AUPR} & \textbf{AUROC} \\
            \midrule
             \textbf{Supervised Models} \\
            SPaRCNet \cite{SPaRCNet}       & 0.79M  & 78.96 $\pm$ 0.18 & 0.8414 $\pm$ 0.0018 & 0.8676 $\pm$ 0.0012 \\
            CNN-Transformer \cite{CNN-Transformer}  & 3.2M   & 77.77 $\pm$ 0.22 & 0.8433 $\pm$ 0.0039 & 0.8461 $\pm$ 0.0013 \\
            FFCL \cite{FFCL}             & 2.4M   & 78.48 $\pm$ 0.38 & 0.8448 $\pm$ 0.0065 & 0.8569 $\pm$ 0.0051 \\
            ST-Transformer \cite{ST-Transformer}  & 3.5M   & 79.66 $\pm$ 0.23 & 0.8521 $\pm$ 0.0026 & 0.8707 $\pm$ 0.0019 \\
            \midrule
            \textbf{Self-supervised Models} \\
            BrainBERT \cite{BrainBERT}  & 43.18M & - & 0.8460 $\pm$ 0.0030 & 0.8530 $\pm$ 0.0020 \\
            EEGFormer Small \cite{EEGFormer} & -   & - & 0.8620 $\pm$ 0.0050 & 0.8620 $\pm$ 0.0070 \\
            EEGFormer Base \cite{EEGFormer} & -   & - & 0.8670 $\pm$ 0.0020 & 0.8650 $\pm$ 0.0010 \\
            EEGFormer Large \cite{EEGFormer} & -   & - & 0.8720 $\pm$ 0.0010 & 0.8760 $\pm$ 0.0030 \\
            LaBraM \cite{LaBraM} & 5.8M   & 81.40 $\pm$ 0.19 & 0.8965 $\pm$ 0.0016 & 0.9022 $\pm$ 0.0009 \\
            LaBraM \cite{LaBraM} & 46M    & 82.26 $\pm$ 0.15 & 0.9130 $\pm$ 0.0005 & 0.9127 $\pm$ 0.0005 \\
            LaBraM \cite{LaBraM} & 369M   & \textbf{82.58} $\pm$ 0.11 & \textbf{0.9204} $\pm$ 0.0011 & \textbf{0.9162} $\pm$ 0.0016 \\
            \midrule
            \textbf{CEReBrO} & 3.58 M & 79.40 $\pm$ 0.19  & 0.8763 $\pm$ 0.0031  & 0.8749 $\pm$ 0.0033 \\
            \textbf{CEReBrO} & 39.95 M &  81.29 $\pm$ 0.15  & 0.8994 $\pm$ 0.0002  & 0.8867 $\pm$ 0.0006 \\
            \textbf{CEReBrO} & 85.15 M & 81.67 $\pm$ 0.23  & 0.9049 $\pm$ 0.0026  & 0.8916 $\pm$ 0.0038 \\
            \bottomrule
        \end{tabular}%
    }
    \caption{Performance on anomaly classification using TUAB.}
    \vspace{-4mm}
    \label{tab:tuab_comparison}
\end{table}

Finally, \cref{tab:tuab_comparison} compares the performance of our approach to state-of-the-art in anomaly classification tasks. Among the supervised models, ST-Transformer achieves the highest accuracy ($79.66\%$) with a model size of 3.5M. LaBraM surpasses ST-Transformer in accuracy by approx. $+2\%$ when keeping a comparable model size and by $+3\%$ when considering a much larger model (369M).
In comparison, \textbf{\gls{cerebro}} achieves a performance comparable to ST-Transformer at a similar model size. Furthermore, we achieve a maximum accuracy comparable to LaBraM with a model size up to $4\times$ smaller. A detailed comparison of LaBraM and \gls{cerebro} can be found in \cref{appendix:labram_comparisons}.

\section{Conclusion}

We propose \gls{cerebro}, a compact encoder-only model with a novel alternating attention mechanism. Our model efficiently captures the spatio-temporal characteristics of \gls{eeg} signals. For 3 model sizes between 3.6 million and 85 million parameters, alternating attention achieves 2$\times$ the speed improvements and consumes 6$\times$ less memory than standard self-attention especially for long input sequences. \gls{cerebro} is pre-trained on public heterogenous \gls{eeg} signals and displays impressive performance on public downstream benchmarks, often matching or surpassing larger and more resource-intense methods. Although EEG foundation models have not yet reached the scale of modern \glspl{llm}, there is substantial potential in developing smaller,  more efficient architectures tailored to EEG data.

\section{Acknowledgements}
This work was supported by the ETH Future Computing Laboratory (EFCL), financed by a donation from Huawei Technologies. This work was supported by a grant from the Swiss National Supercomputing Centre (CSCS) under project ID lp12 on Alps.

{
    \small
    \bibliographystyle{ieeenat_fullname}
    \bibliography{main}
}

\clearpage
\maketitlesupplementary

\appendix

\section{Experiment Details}
\label{appendix:experiment_details}

\subsection{Dataset Pre-processing}

\begin{figure}[ht]
    \centering
    \resizebox{\columnwidth}{!}{ 
        \includegraphics{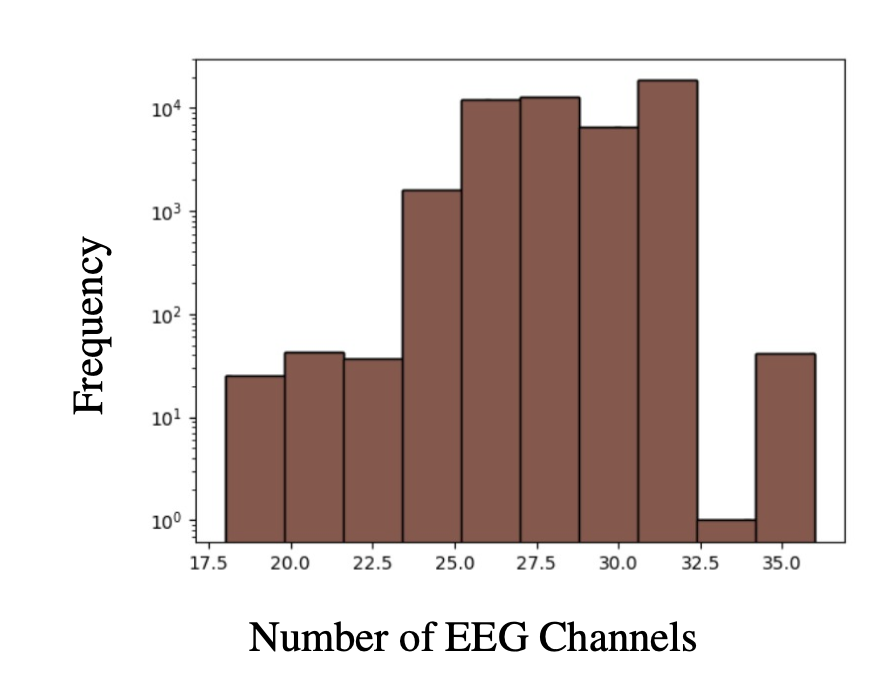}  
    }
    \caption{Channel configurations per .edf File in the TUEG Dataset.}
    \label{fig:tueg_channel_histogram}
\end{figure}

This section describes the datasets used in each of our experiments. In \cref{subsec:waveform_vs_spectrograms}, \cref{subsec:patching_experiments} and \cref{subsec:attention_experiments} we use TUSL  \cite{obeid2016temple}, TUSZ \cite{obeid2016temple} and TUEP \cite{obeid2016temple} for pre-training and TUAB for fine-tuning. The smaller pre-training corpus is selected to align with the reduced computational demands of conducting focused ablation studies. In \cref{subsec:sota_comparisons}, we use TUEG (with the TUAB subset excluded) for pre-training. Note that TUSL, TUSZ and TUEP are subsets of TUEG. We use TUAB, Neonate and SEED for fine-tuning. We incorporate an additional downstream task by fine-tuning \gls{cerebro} on MoBI (see \cref{{appendix:additional_downstream_tasks}}). Dataset details can be found in \cref{tab:dataset_details}. TUEG contains various channel configurations, illustrated in \cref{fig:tueg_channel_histogram}.

\begin{table*}[ht]
\centering
\resizebox{\textwidth}{!}{  
\begin{tabular}{lcccc}
\toprule
\textbf{Dataset} & \textbf{\# Samples} & \textbf{\# Channels} & \textbf{Sample Duration (s)} & \textbf{Sampling Frequency (Hz)} \\
\midrule
TUEP     & 303,081 & 23  & 5   & 256 \\
TUSL     & 16,503  & 23  & 5   & 256 \\
TUSZ     & 598,944 & 23  & 5   & 256 \\
TUAB     & 819,561 & 23  & 5   & 256 \\
Neonate  & 80,536  & 19  & 5   & 256 \\
MoBI     & 542,444 & 60 & 2 & 100 \\
SEED     & 152,730 & 62 & 1 & 200 \\
TUEG (w/o TUAB subjects) & 15,783,482 & 18-36 & 5 & 256 \\
\bottomrule
\end{tabular}%
}
\caption{Dataset details.}
\label{tab:dataset_details}
\end{table*}

\subsection{Training Hyperparameters}
This section details hyperparameters used to pre-train and fine-tune the models discussed in \cref{subsec:sota_comparisons}. Our pre-training hyperparameters are available in \cref{tab:pretraining_hyperparams}. Our fine-tuning hyperparameters are available in \cref{tab:finetune_hyperparams}.

\begin{table*}[ht]
    \centering
    \resizebox{\textwidth}{!}{
    \begin{tabular}{lccc}
    \toprule
    & \textbf{\gls{cerebro} Small} & \textbf{\gls{cerebro} Base} & \textbf{\gls{cerebro} Large} \\
    \midrule
    \textbf{Batch Size} & \multicolumn{3}{c}{4096} \\
    \textbf{Scheduler} & \multicolumn{3}{c}{Cosine + Linear warmup} \\
    \textbf{Optimizer} & \multicolumn{3}{c}{AdamW} \\
    \textbf{Betas} & \multicolumn{3}{c}{[0.9, 0.98]} \\
    \textbf{Peak learning rate} & \multicolumn{3}{c}{1.25e-3} \\
    \textbf{Minimum learning rate} & \multicolumn{3}{c}{2.5e-7} \\
    \textbf{Maximum allowed epochs} & \multicolumn{3}{c}{100} \\
    \textbf{Training stopped at epoch} & \multicolumn{3}{c}{30} \\
    \textbf{Warmup epochs} & \multicolumn{3}{c}{3} \\
    \textbf{Masking ratio} & \multicolumn{3}{c}{0.5} \\
    \textbf{Encoder layers} & 8 & 10 & 12 \\
    \textbf{Number of attention heads per layer} & \multicolumn{3}{c}{12} \\
    \textbf{Embedding dimension} & 192 & 576 & 768 \\
    \textbf{MLP Size} & 768 & 2304 & 3072 \\
    \textbf{Weight decay} & \multicolumn{3}{c}{0.05} \\
    \textbf{Patch size} & \multicolumn{3}{c}{64} \\
    \bottomrule
    \end{tabular}%
    }
    \caption{Pre-training hyperparameters. The same hyperparameters are used across all model sizes, with a few exceptions specified in this table.}
    \label{tab:pretraining_hyperparams}
\end{table*}

\begin{table*}[ht]
    \centering
    \resizebox{\textwidth}{!}{ 
    \begin{tabular}{lccc}
    \toprule
    & \textbf{\gls{cerebro} Small} & \textbf{\gls{cerebro} Base} & \textbf{\gls{cerebro} Large} \\
    \midrule
    \textbf{Batch Size} & \multicolumn{3}{c}{4096} \\
    \textbf{Scheduler} & \multicolumn{3}{c}{Cosine + Linear warmup} \\
    \textbf{Optimizer} & \multicolumn{3}{c}{AdamW} \\
    \textbf{Betas} & \multicolumn{3}{c}{[0.9, 0.999]} \\
    \textbf{Peak learning rate} & \multicolumn{3}{c}{5e-4} \\
    \textbf{Minimum learning rate} & \multicolumn{3}{c}{2.5e-7} \\
    \textbf{Total epochs} & \multicolumn{3}{c}{50} \\
    \textbf{Warmup epochs} & \multicolumn{3}{c}{5} \\
    \textbf{Layer-wise learning rate decay factor} & \multicolumn{3}{c}{0.75} \\
    \textbf{Weight decay} & \multicolumn{3}{c}{0.05} \\
    \textbf{Gaussian noise amplitude ratio} & \multicolumn{3}{c}{0.2} \\
    \textbf{Gaussian noise addition probability} & \multicolumn{3}{c}{0.5} \\
    \textbf{Drop path} & 0.1 & 0.2 & 0.2\\
    \textbf{Label smoothing} & \multicolumn{3}{c}{0.1} \\
    \bottomrule
    \end{tabular}%
    }
    \caption{Fine-tuning hyperparameters. The same fine-tuning hyperparameters are used for all three model sizes, with a few exceptions specified in this table.}
    \label{tab:finetune_hyperparams}
\end{table*}

\section{Alternatives to Self-Attention}
\label{appendix:split_attention}

Our tokenization scheme (see \cref{subsec:tokenization}) generates thousands of tokens per training example. Standard self-attention has quadratic memory and time complexity in the input sequence length, which, given our tokenization, limits the potential for deployment in resource-constrained environments. This section discusses the alternatives to self-attention that were explored to address this challenge. We finally selected alternating attention by comparing runtime, memory requirements, and impact on downstream performance (see \cref{alg:alternating_attention}).

\subsection{Two-Axis Attention}
In two-axis attention (\cref{alg:twoaxis_attention}), we compute intra-channel and inter-channel attention in each encoder block using two separate QKV projections. Unlike in alternating attention, each encoder block contributes to learning the spatio-temporal characteristics of the input EEG signals. The primary limitation is that the additional QKV projection in each encoder layer adds more parameters and, therefore, increases the complexity of the model. For example, in \gls{cerebro} Large, two-axis attention yields approximately 20M additional trainable parameters compared to alternating attention.

\begin{algorithm}
\caption{\textproc{Two-Axis Attention}}
\begin{algorithmic}[1]
\Require Input tensor $\mathbf{T}$ of shape $[B, C \times N_p, d_e]$
\Statex \textbf{Parameters:}
\Statex \quad $B$: batch size
\Statex \quad $C$: number of channels
\Statex \quad $N_p$: number of patches per channel
\Statex \quad $d_e$: embedding dimension
\For{each attention block}
    \State Compute projections: $Q_c, K_c, V_c$ and $Q_p, K_p, V_p$
    \State Reshape $\mathbf{T}$ to $[B \times N_p, C, d_e]$
    \State $\mathbf{A_1}$ = Multi-head attention over channels using $(Q_c, K_c, V_c)$
    \State Reshape $\mathbf{T}$ to $[B \times C, N_p, d_e]$
    \State $\mathbf{A_2}$ = Multi-head attention over patches using $(Q_p, K_p, V_p)$
    \State Reshape $A_1$ and $A_2$ back to $[B, C, N_p, d_e]$
    \State Compute $\text{Mean}(\mathbf{A_1}, \mathbf{A_2})$ \Comment{Combine outputs}
\EndFor
\Ensure Output tensor of shape $[B, C \times N_p, d_e]$
\end{algorithmic}
\label{alg:twoaxis_attention}
\end{algorithm}

\subsection{Bottleneck Attention}

Bottleneck attention expands upon two-axis attention, intending to have intra-channel and inter-channel attention computation in each encoder block without resorting to two separate QKV projections. The idea is first to compute inter-channel attention and use the resulting attention map as a value vector in the intra-channel attention computation. Details are discussed in \cref{alg:bottleneck_attention}.

\begin{algorithm}
\caption{\textproc{Bottleneck Split Attention}}
\begin{algorithmic}[1]
\Require Input tensor $\mathbf{T}$ of shape $[B, C \times N_p, d_e]$
\Statex \textbf{Parameters:}
\Statex \quad $B$: batch size
\Statex \quad $C$: number of channels
\Statex \quad $N_p$: number of patches per channel
\Statex \quad $d_e$: embedding dimension
\For{each attention block}
    \State Compute QKV projections: $Q$, $K$, $V$ of shape $[B, C, N_p, d_e]$
    \State Compute $(Q, K, V)_{\text{avg}(N_p)} =\text{MeanPool}(Q, K, V)$ over $N_p$, resulting in shape $[B, C, 1, d_e]$
    \State $\mathbf{A_1}=\text{Attention}(Q_{\text{avg}(N_p)}, K_{\text{avg}(N_p)}, V_{\text{avg}(N_p)})$
    \State Compute $(Q, K)_{\text{avg}(C)}=\text{MeanPool}(Q, K)$ over $C$, resulting in shape $[B, 1, N_p, d_e]$
    \State Broadcast $\mathbf{A_1}$ to shape $[B, C, N_p, d_e]$
    \State Compute $\text{Attention}(Q_{\text{avg}(C)}, K_{\text{avg}(C)}, \mathbf{A_1})$
\EndFor
\Ensure Output tensor of shape $[B, C \times N_p, d_e]$
\end{algorithmic}
\label{alg:bottleneck_attention}
\end{algorithm}

\subsection{Theoretical and Empirical Complexity Analysis of Different Attention Mechanisms}

\begin{table}[!ht]
    \centering
    \resizebox{\columnwidth}{!}{%
        \begin{tabular}{lll}
            \toprule
            \textbf{Attention Type} & \textbf{Memory Complexity} & \textbf{Time Complexity} \\
            \midrule
            Alternating (over $p$) & $\mathcal{O}\left( CN_p^2 \right)$ & $\mathcal{O}\left( CN_p^2d_e \right)$ \\
            Alternating (over $C$) & $\mathcal{O}\left( C^2N_p \right)$ & $\mathcal{O}\left( C^2N_pd_e \right)$ \\
            Bottleneck & $\mathcal{O}\left( CN_p(C+N_p) \right)$ & $\mathcal{O}\left( CN_p(C + N_p)d_e \right)$ \\
            Two-Axis & $\mathcal{O}\left( CN_p(C+N_p) \right)$ & $\mathcal{O}\left( CN_p(C + N_p)d_e \right)$ \\
            Standard & $\mathcal{O}\left( (CN_p)^2\right)$ & $\mathcal{O}\left( (CN_p)^2d_e \right)$ \\
            \bottomrule
        \end{tabular}%
    }
    \caption{Theoretical memory and time complexities of each attention mechanism.}
    \label{tab:split_attn_complexity_supp}
\end{table}

\definecolor{green_fig2}{RGB}{0,153,51}
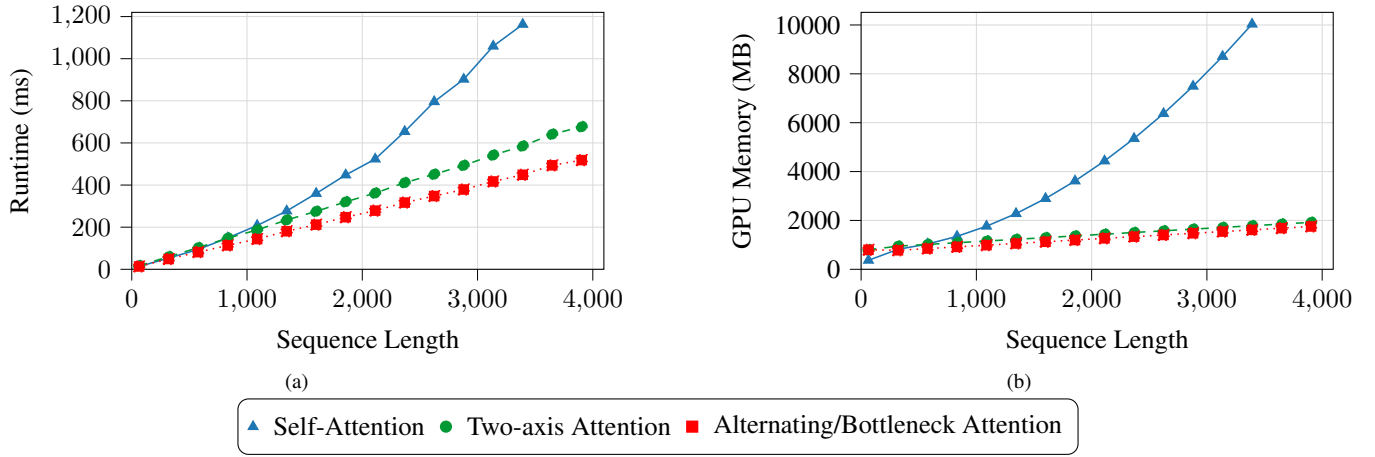
\begin{figure*}[ht]
  \centering
  \begin{subfigure}{0.45\textwidth}
    \centering
    \begin{tikzpicture}
      \begin{axis}[
        width=\textwidth,
        height=5cm,
        tick align=outside,
        tick pos=left,
        xmajorgrids,
        ymajorgrids,
        grid style={line width=.1pt, draw=gray!30},
        xlabel={Sequence Length},
        ylabel={Runtime (ms)},
        xmin=0, xmax=4096,
        ymin=0, ymax=1220,
        xtick style={color=black},
        ytick style={color=black},
        ytick={-200,0,200,400,600,800,1000,1200,1400},
        tick label style={font=\normalsize},
      ]
        \addplot+[semithick, color=majorColor, solid, mark=triangle*, mark size=2]
        table {%
        64 13.4416227340698
        320 53.2770462036133
        576 94.2263870239258
        832 148.674331665039
        1088 207.98193359375
        1344 276.891693115234
        1600 360.196472167969
        1856 447.957824707031
        2112 523.412231445312
        2368 653.839599609375
        2624 795.702026367188
        2880 902.351684570312
        3136 1059.46325683594
        3392 1162.90478515625
        };
        \addplot+[semithick, color=green_fig2, dashed, mark=*, mark size=2]
        table {%
        64 17.0931529998779
        320 60.0352592468262
        576 102.773338317871
        832 147.976654052734
        1088 186.814559936523
        1344 234.259948730469
        1600 275.219604492188
        1856 320.392913818359
        2112 361.959411621094
        2368 411.286926269531
        2624 451.243896484375
        2880 493.347808837891
        3136 542.258666992188
        3392 585.347412109375
        3648 641.520935058594
        3904 676.971984863281
        };
        \addplot+[semithick, color=red, dotted, mark=square*, mark size=2]
        table {%
        64 13.5173635482788
        320 48.9474830627441
        576 81.4892196655273
        832 113.43928527832
        1088 143.637817382812
        1344 180.781356811523
        1600 212.017639160156
        1856 247.271331787109
        2112 278.609558105469
        2368 316.858734130859
        2624 347.806671142578
        2880 378.874206542969
        3136 417.008483886719
        3392 448.833709716797
        3648 493.573699951172
        3904 518.365783691406
        };
      \end{axis}
    \end{tikzpicture}
    \caption{}
    \label{fig:memory2}
  \end{subfigure}
  \hfill
  \begin{subfigure}{0.45\textwidth}
    \centering
    \begin{tikzpicture}
      \begin{axis}[
        width=\textwidth,
        height=5cm,
        tick align=outside,
        tick pos=left,
        xmajorgrids,
        ymajorgrids,
        grid style={line width=.1pt, draw=gray!30},
        xlabel={Sequence Length},
        ylabel={GPU Memory (MB)},
        xmin=0, xmax=4096,
        ymin=0, ymax=10516,
        xtick style={color=black},
        ytick style={color=black},
        ytick={-2000,0,2000,4000,6000,8000,10000,12000},
        yticklabels={-2000,0,2000,4000,6000,8000,10000,12000},
        tick label style={font=\normalsize},
        scaled ticks=false, 
      ]
      \addplot+[semithick, color=majorColor, solid, mark=triangle*, mark size=2]
      table {%
        64 367.987701416016
        320 818.470886230469
        576 1032.73266601562
        832 1347.65771484375
        1088 1763.24609375
        1344 2279.49780273438
        1600 2896.41259765625
        1856 3614.90014648438
        2112 4433.88720703125
        2368 5352.46435546875
        2624 6371.70458984375
        2880 7491.6083984375
        3136 8712.17578125
        3392 10033.4794921875
      };
      \addplot+[semithick, color=green_fig2, dashed, mark=*, mark size=2]
      table {%
        64 797.642761230469
        320 948.330505371094
        576 1017.74951171875
        832 1087.30773925781
        1088 1156.86608886719
        1344 1226.42431640625
        1600 1295.98254394531
        1856 1366.4501953125
        2112 1436.75390625
        2368 1505.98449707031
        2624 1575.21508789062
        2880 1644.44567871094
        3136 1713.67626953125
        3392 1782.98059082031
        3648 1852.44873046875
        3904 1922.00708007812
      };
      \addplot+[semithick, color=red, dotted, mark=square*, mark size=2]
      table {%
        64 797.904907226562
        320 778.100708007812
        576 847.658996582031
        832 917.21728515625
        1088 987.299865722656
        1344 1056.85815429688
        1600 1126.41638183594
        1856 1196.88391113281
        2112 1267.18774414062
        2368 1336.41833496094
        2624 1405.64892578125
        2880 1474.87951660156
        3136 1544.11010742188
        3392 1613.41442871094
        3648 1682.88256835938
        3904 1752.44079589844
      };
      \end{axis}
    \end{tikzpicture}
    \caption{}
    \label{fig:runtime}
  \end{subfigure}
  \vspace{0.1em}
  \begin{tikzpicture}
    \node[draw=black, rounded corners, inner sep=2pt] (legend) {
      \begin{tikzpicture}[font=\normalsize]
        \matrix [column sep=4pt, row sep=2pt] {
          \draw[semithick, color=majorColor, solid, mark=triangle*, mark options={solid}, mark size=2]
            plot coordinates {(0,0)}; &
          \node[anchor=west] {Self-Attention}; &
          \draw[semithick, color=green_fig2, dashed, mark=*, mark options={solid}, mark size=2]
            plot coordinates {(0,0)}; &
          \node[anchor=west] {Two-axis Attention}; &
          \draw[semithick, color=red, dotted, mark=square*, mark options={solid}, mark size=2]
            plot coordinates {(0,0)}; &
          \node[anchor=west] {Alternating/Bottleneck Attention}; \\
        };
      \end{tikzpicture}
    };
  \end{tikzpicture}

  \caption{(a) Runtime vs input sequence length for different attention mechanisms. (b) GPU memory usage vs. input sequence length for different attention mechanisms. In these experiments we use \gls{cerebro} Large on 4 NVIDIA RTX 2080 Ti GPUs. We use $N_p = 20$ and $C \in [1, 64]$ to replicate typical \gls{eeg} configurations. For self-attention, we run out of GPU memory for sequence lengths larger than 3392.}
  \label{fig:split_attn_comparisons}
\end{figure*}

In \cref{fig:split_attn_comparisons}, we plot runtime and GPU memory consumption of our different attention mechanisms as functions of input sequence lengths. Several key patterns emerge. First, it is clear that two-axis, bottleneck, and alternating attention all require less memory (up to $5 \times$ less) and have quicker runtimes (more than $2\times$ faster) than standard self-attention. Bottleneck attention and alternating attention have very similar memory and runtime profiles. The benefits of bottleneck, two-axis, and alternating attention are highlighted for large sequence lengths. Our findings in \cref{tab:appendix_attention_comparison} corroborate this, which suggests that the true benefit emerges when $C + N_p << CN_p$.

\subsection{Performance Comparisons for Different Attention Mechanisms}

\begin{table}[!ht]
    \centering
    \resizebox{\columnwidth}{!}{%
        \begin{tabular}{llccc}
            \toprule
            \textbf{Representation} & \textbf{Attention} & \textbf{Balanced Accuracy (\%)} & \textbf{AUPR} & \textbf{AUROC} \\
            \midrule
            \multirow{4}{*}{Spectrogram} 
             & Bottleneck  & 77.49 & 0.8548 & 0.8545 \\
             & Alternating & 78.49 & 0.8684 & 0.8674 \\
             & Two-Axis    & 78.83 & 0.8702 & 0.8671 \\
             & Standard    & 78.89 & 0.8719 & 0.8690 \\
            \midrule
            \multirow{4}{*}{Waveform} 
             & Bottleneck  & 78.20 & 0.8640 & 0.8594 \\
             & Standard    & 79.35 & 0.8755 & 0.8700 \\
             & Two-Axis    & 80.57 & 0.8885 & 0.8820 \\
             & Alternating & \textbf{80.65} & \textbf{0.8901} & \textbf{0.8825} \\
            \bottomrule
        \end{tabular}%
    }
    \caption{The effect of each attention mechanism in pre-training and fine-tuning on anomaly classification.}
    \label{tab:appendix_attention_comparison}
\end{table}

In \cref{tab:appendix_attention_comparison}, we observe distinct spectrogram results compared to waveforms. Standard self-attention yields the most competitive results across all three spectrogram metrics, closely followed by two-axis and alternating attention. In contrast, two-axis and alternating attention outperform standard self-attention for waveforms. In both regimes, bottleneck split attention is considerably worse than standard self-attention. We also observe that the best waveform representation yields significant improvements (notably $+ 1.75\%$ in the balanced accuracy, $+ 0.0182$ in the AUPR and $+ 0.0135$ in the AUROC) compared to the best spectrogram representation. A waveform representation yields better performance across all metrics than a spectrogram representation. Based on these results, our final model (presented in \cref{sec:methodology} and \cref{sec:experiments}) uses waveform representations of EEG signals and alternating attention. This combination yields the best balance between efficiency (as presented in \cref{fig:split_attn_comparisons}) and performance (as presented in \cref{tab:appendix_attention_comparison}). 

\section{Incorporating Magnitude and Phase Information Into Spectrogram Tokenization} \label{appendix:spectrogram_ablations}

In \cref{tab:mag_vs_phase}, we study the impact of combining magnitude and phase information when spectrograms are used as input to our model. Specifically, we introduce phases as additional "channels" in our spectrogram tensors, and we evaluate the resulting performance on the TUAB dataset.

We follow two approaches to introduce the phase: the interleaved approach and the separated approach.

In the interleaved approach, we construct the channels as
\begin{equation}
    \resizebox{\columnwidth}{!}{$
        [\text{magnitude}_1, \text{ phase}_1, \text{ magnitude}_2, \text{ phase}_2, \dots, \text{ magnitude}_{C}, \text{ phase}_{C}]
    $}
\end{equation}

where magnitude$_i$ and phase$_i, i \in [1, 23]$ represent the pre-processed magnitude and phase tensors of channel $i$.

In the separated approach, we construct the channels as
\begin{equation}
    \resizebox{\columnwidth}{!}{$
        [\text{magnitude}_1, \text{ magnitude}_2, \dots, \text{ magnitude}_{23}, \text{ phase}_1, \text{ phase}_2, \dots, \text{ phase}_{23}]
    $}
\end{equation}

Including phase information doubles the number of tokens in each training example, and we aim to evaluate if this addition results in an increased accuracy.

\cref{tab:mag_vs_phase} illustrates the performance of \gls{cerebro} Large trained from scratch on TUAB with and without phase information. No statistical evidence suggests that including phase information, whether in the interleaved or separated regime, yields a performance improvement.

\begin{table}[ht]
    \centering
    \resizebox{\columnwidth}{!}{
        \begin{tabular}{l c}
            \toprule
            \textbf{Spectrogram Information} & \textbf{Balanced Accuracy (\%)} \\
            \midrule
            Magnitude only & 77.01 \\
            Magnitude + interleaved phase & 76.01 \\
            Magnitude + separated phase & 76.89 \\
            \bottomrule
        \end{tabular}
    }
    \caption{Comparison of spectrogram magnitude vs. magnitude and phase information on anomaly classification.}
    \label{tab:mag_vs_phase}
\end{table}

\section{Additional Downstream Tasks}
\label{appendix:additional_downstream_tasks}

In this section, we provide additional downstream tasks. In particular, we use the MoBI dataset \cite{mobi}. Mobile brain-body imaging data is collected during a treadmill-based BCI task involving lower limb motor imagery. Six goniometers recorded bilateral joint angles (hip, knee, and ankle), aiming to regress 12 targets (left and right leg angles). Data were gathered from 8 healthy subjects, each completing three identical trials. During fine-tuning, we use an MSE loss function and report $R^2$ and RMSE scores as performance metrics. Results are included in \cref{tab:mobi_performance}.

\begin{table}[!ht]
    \centering
    \resizebox{\columnwidth}{!}{%
        \begin{tabular}{lccc}
            \toprule
            \textbf{Methods} & \textbf{Model Size} & \textbf{$\textbf{R}^2$ Score} & \textbf{RMSE} \\
            \midrule
            \textbf{Supervised Models} \\
            SPaRCNet \cite{SPaRCNet}       & 0.79M & 0.1467 $\pm$ 0.0064 & 0.1344 $\pm$ 0.0006 \\
            CNN-Transformer \cite{CNN-Transformer} & 3.2M & 0.0628 $\pm$ 0.0089 & 0.1411 $\pm$ 0.0007 \\
            FFCL \cite{FFCL}               & 2.4M & 0.0712 $\pm$ 0.0124 & 0.1396 $\pm$ 0.0014 \\
            ST-Transformer \cite{ST-Transformer}  & 3.5M & 0.2911 $\pm$ 0.0014 & 0.1222 $\pm$ 0.0001 \\
            \midrule
            \textbf{Self-supervised Models} \\
            BIOT \cite{BIOT}               & 3.2M & 0.0597 $\pm$ 0.0069 & 0.1401 $\pm$ 0.0006 \\
            LaBraM-Base \cite{LaBraM}      & 5.8M & 0.2876 $\pm$ 0.0032 & 0.1225 $\pm$ 0.0003 \\
            LaBraM-Large \cite{LaBraM}     & 46M & 0.3093 $\pm$ 0.0032 & 0.1197 $\pm$ 0.0003 \\
            LaBraM-Huge \cite{LaBraM}      & 369M & \textbf{0.3145} $\pm$ 0.0032 & \textbf{0.1196} $\pm$ 0.0003 \\
            \midrule
            \textbf{CEReBrO} & 3.58M & 0.1036 $\pm$ 0.0007 & 0.1375 $\pm$ 0.0005\\
            \textbf{CEReBrO} & 39.95M & 0.2776 $\pm$ 0.0002 & 0.1239 $\pm$ 0.0006 \\
            \textbf{CEReBrO} & 85.15M & 0.3118 $\pm$ 0.0032 & 0.1209 $\pm$ 0.0004 \\
            \bottomrule
        \end{tabular}%
    }
    \caption{Performance on gait prediction using MoBI.}
    \label{tab:mobi_performance}
\end{table}

In \cref{tab:mobi_performance}, we see that \gls{cerebro} Large surpasses LaBraM Large and LaBraM base in the $R^2$ score while yielding comparable RMSE. However, it performs slightly worse than LaBraM Huge, which is $4\times$ larger. Overall, even in a regression task, we see that \gls{cerebro} remains competitive compared to the current state of the art.

\section{Comparing LaBraM and \gls{cerebro}}
\label{appendix:labram_comparisons}

LaBraM yields competitive results at all model sizes, frequently surpassing \gls{cerebro}. However, we posit that several aspects of LaBraM make it challenging to train and deploy in resource-constrained environments. Firstly, LaBraM requires multi-stage pre-training: a discrete neural tokenizer has to be pre-trained, followed by the LaBraM encoder. This increases LaBraM's memory footprint and pre-training time,  making LaBraM less flexible in the face of distribution shifts downstream. LaBraM's discrete neural tokenizer also has limitations. For example, codebook size selection is non-trivial and highly depends on the characteristics of the input data. Secondly, discrete codebooks have been shown to suffer from information loss \cite{li2024continuousspeechtokenizertext}. LaBraM's encoder uses standard self-attention, which may capture more correlations in the data, but is also more memory-intensive and slower than alternating attention, especially for larger sequence lengths. In \cref{tab:cerebro_vs_labram}, we illustrate the peak memory usage of LaBraM and \gls{cerebro}. While the memory usage is comparable, these values do not include the non-trivial memory usage of LaBraM's neural tokenizer, especially during pre-training. 

\begin{table}[!ht]
    \centering
    \resizebox{\columnwidth}{!}{%
    \begin{tabular}{lccc}
        \toprule
        \textbf{Model} & \textbf{Model Size} & \textbf{Peak GPU Memory Usage (MB)} \\
        \midrule
        \multirow{3}{*}{LaBraM} 
            & 5.8M & 757.38 \\
            & 46M  & 1371.92 \\
            & 369M & 2758.42 \\
        \midrule
        \multirow{3}{*}{CEReBrO} 
            & 3.58M  & 402.96 \\
            & 39.95M & 1016.31 \\
            & 85.15M & 1620.29 \\
        \bottomrule
    \end{tabular}}
    \caption{Comparison of Peak GPU Memory Usage for the best performing LaBraM and \gls{cerebro} models. In each model, we use $C = 64$ and batch size $= 8$.}
    \label{tab:cerebro_vs_labram}
\end{table}

\printglossary

\end{document}